\def\year{2018}\relax
\documentclass[letterpaper]{article} 
\usepackage{aaai18}  
\usepackage{times}  
\usepackage{helvet}  
\usepackage{courier}  
\usepackage{url}  
\usepackage{graphicx}  
\usepackage{amsmath}
\usepackage{amsfonts}
\usepackage{amssymb}
\usepackage{mathrsfs}
\usepackage{amsthm}
\usepackage{amsmath}
\usepackage{bm}
\newtheorem{defi}{Definition}
\DeclareMathOperator*{\argmax}{arg\,max}
\usepackage{algorithm}	
\usepackage{algorithmic}
\usepackage{color}
\usepackage{multirow}
\usepackage{graphicx}
\usepackage{subfigure}
\usepackage{fixltx2e}

\begin{document}
%

\title{Domain Generalization via Conditional Invariant Representations}
\author{Ya Li${^\dag}$, Mingming Gong${^{\bigstar \divideontimes}}$, Xinmei Tian${^\dag}$, Tongliang Liu${^\ddag}$, Dacheng Tao${^\ddag}$ \\
${^\dag}$CAS Key Laboratory of Technology in Geo-Spatial Information Processing and Application Systems, \\
University of Science and Technology of China, China\\
${^\ddag}$ UBTECH Sydney Artificial Intelligence Institute, SIT, FEIT, The University of Sydney, Australia\\
${^\bigstar}$Department of philosophy, Carnegie Mellon University \\
${^\divideontimes}$Department of Biomedical Informatics, University of Pittsburgh \\
muziyiye@mail.ustc.edu.cn, gongmingnju@gmail.com, xinmei@ustc.edu.cn,  \\
tliang.liu@gmail.com, dacheng.tao@sydney.edu.au
}
\maketitle
\begin{abstract}
Domain generalization aims to apply knowledge gained from multiple labeled source domains to unseen target domains. The main difficulty comes from the dataset bias: training data and test data have different distributions, and the training set contains heterogeneous samples from different distributions. Let $X$ denote the features, and $Y$ be the class labels. Existing domain generalization methods address the dataset bias problem by learning a domain-invariant representation $h(X)$ that has the same marginal distribution $\mathbb{P}(h(X))$ across multiple source domains. The functional relationship encoded in $\mathbb{P}(Y|X)$ is usually assumed to be stable across domains such that $\mathbb{P}(Y|h(X))$ is also invariant. However, it is unclear whether this assumption holds in practical problems. In this paper, we consider the general situation where both $\mathbb{P}(X)$ and $\mathbb{P}(Y|X)$ can change across all domains. We propose to learn a feature representation which has domain-invariant class conditional distributions $\mathbb{P}(h(X)|Y)$. With the conditional invariant representation, the invariance of the joint distribution $\mathbb{P}(h(X),Y)$ can be guaranteed if the class prior $\mathbb{P}(Y)$ does not change across training and test domains. Extensive experiments on both synthetic and real data demonstrate the effectiveness of the proposed method.
\end{abstract}

\section{Introduction}
Recent years have witnessed a great success of supervised learning in various pattern recognition problems, such as image classification, object detection, and speech recognition. Standard supervised learning relies heavily on the {\it i.i.d.} data assumption; however, dataset-bias is unavoidable in many situations due to selection bias or mechanism changes. For example, this problem has been well recognized in the computer vision community \cite{torralba2011unbiased,khosla2012undoing}: the widely adopted vision datasets have their special properties and are not representative of the visual world. In medical diagnosis, the distribution of cell types varies from patient to patient, and we need to train a classifier on the data collected from previous patients that generalizes well to unseen patients \cite{blanchard2011generalizing,muandet9}. These problems are known as domain generalization, in which the training set consists of data from heterogeneous source domains, say patients, and the test data distribution is different from that of the training data. 

To handle the distribution changes, many existing domain generalization methods aim to learn domain-invariant representations that have stable distributions across all source domains \cite{muandet9,erfani23,ghifary8}. The learned invariant representations are expected to generalize well to any unseen test set under the assumption that the changes of distribution across source and test domains are caused by some common factors whose effects are removed in the invariant representations. In computer vision, such factors could be illumination, camera viewpoints,  and backgrounds. These methods have achieved good performance in computer vision \cite{ghifary15,ghifary8} and medical diagnosis \cite{muandet9}.  

However, existing methods that learn domain-invariant representations assume that only $\mathbb{P}(X)$ changes across domains while the conditional distribution $\mathbb{P}(Y|X)$ is rather stable. Thus, the conditional distribution $\mathbb{P}(Y|h(X))$ is also invariant, and the learning problem reduces to ensuring that the marginal distribution $\mathbb{P}(h(X))$ is invariant across domains. This assumption greatly simplifies the problem, but it is unclear whether this assumption holds in practical situations. According to some recent results in causal learning \cite{scholkopf16,janzing17}, $\mathbb{P}(Y|X)$ can be stable when $\mathbb{P}(X)$ changes in the situation where $X$ is the cause for $Y$, i.e., the causal structure is $X\rightarrow Y$. This is because the mechanism that generates the cause, i.e., $\mathbb{P}(X)$, is not coupled with the mechanism that generates the effect from the cause, i.e., $\mathbb{P}(Y|X)$, and not vice versa. That is to say, if $Y$ is the cause and $X$ is the effect, $\mathbb{P}(X)$ often changes together with $\mathbb{P}(Y|X)$. In this situation, if $\mathbb{P}(X)$ changes, it is very likely that $\mathbb{P}(Y|X)$ also changes across domains, which violates the stability of $\mathbb{P}(Y|X)$ assumption. In practice, we have plenty of problems where the causal structure is $Y\rightarrow X$. For example, , in face recognition, Y is person id, X is the feature, and $\theta$ is the viewpoint. Let us consider each viewpoint as a domain, then in each domain we have conditional distribution $P(X|Y,\theta=\theta_i)$. According to Bayes theorem, $P(Y|X,\theta=\theta_i) = P(X|Y,\theta=\theta_i)P(Y|\theta=\theta_i)/P(X|\theta=\theta_i),$ thus changes across domains. This conflicts with previous assumptions that $P(Y|X)$ keeps unchanged. There are also other examples, e.g. speaker recognition and person re-identification \cite{yang2017enhancing}.

In this paper, we assume both $\mathbb{P}(X)$ and $\mathbb{P}(Y|X)$ change across domains. We aim to find a feature transformation $h(X)$ that has invariant class-conditional distribution $\mathbb{P}(h(X)|Y)$. To achieve so, we propose to minimize two regularization terms that enforce distribution invariance across source domains. The first term measures the variance of each class-conditional distribution across all source domains and then sums up the variances for all classes. The second term is the variance of class prior-normalized marginal distribution $\mathbb{P}_N(h(X))$, which measures the global distribution discrepancy. The normalization of class priors is introduced to remove the effects brought by possible changes in $\mathbb{P}(Y)$ across source domains. If the prior distribution $\mathbb{P}(Y)$ does not change across source domains, the second term reduces to the common technique used in existing domain-invariant representation learning methods \cite{muandet9,ghifary8}. To preserve the discriminative power of the learned representation, we also incorporate the intra-class and inter-class distances used in kernel Fisher discriminant analysis (FDA)\cite{mika1999fisher}. \par

Compared to existing domain-invariant representation learning methods, our method does not require the assumption of stable $\mathbb{P}(Y|X)$ by exploiting the labels on the source domains which were overlooked in the previous methods. Especially, if the prior distribution $\mathbb{P}(Y)$ on the test sets is the same as that on the training set containing all source domains, our method is able to learn representations $h(X)$ that have invariant joint distribution $\mathbb{P}(h(X),Y)$ across all domains. We conduct a series of experiments on both synthetic and real data, and the results demonstrate the effectiveness of our method.

\section{Related Work}
Domain generalization has been widely applied in classification tasks \cite{xu13,duan14,muandet9,ghifary8,ghifary15,erfani23}. Compared with standard supervised learning, domain generalization methods aim to reduce data bias across different domains and improve the generalization of the learned model to unseen but related domains. For example, \cite{xu13}assumed that positive samples within the same shared latent domain should have similar likelihood and proposed to exploit the low-rank structure from latent domains for domain generalization. \cite{muandet9} proposed domain-invariant component analysis (DICA) through learning an invariant feature representation $h(X)$, in which the difference between marginal distributions $\mathbb{P}(h(X))$ is minimized. \cite{ghifary8} proposed a unified framework called scatter component analysis for domain adaptation and domain generalization. The scatter component analysis combines domain scatter \cite{muandet9}, kernel PCA \cite{scholkopf1998nonlinear}, and kernel FDA \cite{mika1999fisher} in a single objective function. However, all these methods assume that the distribution between domains differs only in the marginal distribution $\mathbb{P}(X)$ while the conditional distribution $\mathbb{P}(Y|X)$ keeps stable or unchanged across domains.  This assumption can simplify the problem of domain generalization, but it is easily violated in real-world applications. \par

Domain adaptation is a related problem which has been extensively studied in the literature \cite{6751205,Huang07,Pan11,pmlr-v70-long17a,shao2014generalized,shao2016spectral,luogeneral,liu2017understanding,}. Assuming that only $\mathbb{P}(X)$ changes, the distribution changes can be corrected by importance reweighting \cite{Huang07} or domain-invariant feature learning \cite{Pan11,6751205}, using unlabeled data from source and target domains. Recently, several works attempted to work in the situation where both $\mathbb{P}(X)$ and $\mathbb{P}(Y|X)$ change across domains \cite{zhang11,gong10,pmlr-v70-long17a}. \cite{zhang11} and \cite{gong10} proposed to consider the domain adaptation problem in the generalized target shift (GeTarS) scenario where the causal direction is $Y \rightarrow X$. In this scenario, both the change of distribution $\mathbb{P}(Y)$ and conditional distribution $\mathbb{P}(X|Y)$ are considered to reduce the data bias across domains. \cite{zhang11} made an assumption that features from source domains can be transferred to the target domain by a location-scale transformation, which is restricted in real-world applications because of the presence of noises in features. \cite{gong10} proposed to le{}arn components whose conditional distribution $\mathbb{P}(h(X)|Y)$ is invariant across domains and estimate the target label distribution $\mathbb{P}^t(Y)$ through labeled source domain data and unlabeled target domain data. Since there are no labels in the target domain to match class-conditionals, the invariance of $\mathbb{P}(h(X)|Y)$ is achieved by minimizing the discrepancy of the marginal distribution $\mathbb{P}(h(X))$ under some untestable assumptions. \cite{pmlr-v70-long17a} proposed an iterative way to match the conditionals by using the predicted labels from previous iterations as pseudo labels. Different from the domain adaptation methods, domain generalization does not require unlabeled data from the target domains.

\section{Conditional Invariant Domain Generalization}
In this section, we first establish the basic notations of domains and formally introduce the definition of domain generalization. Then we give a detailed description of the proposed conditional invariant domain generalization (CIDG) method. 

\subsection{Problem Definition}
Denote $\mathcal{X}$ and $\mathcal{Y}$ as the input feature and label spaces, respectively. A domain defined on $\mathcal{X} \times \mathcal{Y}$ can be represented by a joint probability distribution $\mathbb{P}(X,Y)$. For simplicity, we denote the joint probability distribution $\mathbb{P}^s(X,Y)$ of the $s$-th source domain as $\mathbb{P}^s$. The domain $\mathbb{P}^s$ is associated with a sample $D_s = \{x_i^s, y_i^s\}_{i=1}^{n^s}$, where $(x_i^s,y_i^s) \sim \mathbb{P}^s$ and $n^s$ denotes the sample size of the domain $\mathbb{P}^s$. Then we can define domain generalization as follows. \par

\begin{defi}[Domain Generalization]
Given multiple related source domains $\Omega = \{\mathbb{P}^1,\mathbb{P}^2,...,\mathbb{P}^m\}$ and each domain is associated with a sample $D_s=\{x_i^s,y_i^s\}_{i=1}^{n^s} \sim \mathbb{P}^s$, where $s=\{1,2,\ldots,m\}$. The goal of domain generalization is to learn a classification function $f: \mathcal{X} \rightarrow \mathcal{Y}$ from source domain datasets $\{D_s\}_{s=1}^m$ and apply it to an unseen but related target domain $\mathbb{P}^t(X,Y)$. 
\end{defi}



\subsection{Kernel Mean Embedding}
Before introducing the proposed method, we briefly review the kernel mean embedding of distributions, which is an important mathematical tool to represent and compare distributions \cite{song2013kernel,sriperumbudur2010hilbert}. Let $\mathcal{H}$ denote a characteristic reproducing kernel Hilbert space (RKHS) on $\mathcal{X}$ associated with a kernel $k(\cdot,\cdot): \mathcal{X} \times \mathcal{X} \rightarrow \mathbb{R}$, and $\phi$ be an associated mapping such that $\phi(x) \in \mathcal{H}$. Suppose we have two observations $x_1^s \in \mathcal{X}$ and $x_2^s \in \mathcal{X}$ from domain $s$, then we have $\left<\phi(x_1^s),\phi(x_2^s)\right> = k(x_1^s,x_2^s)$. The kernel embedding of a distribution $\mathbb{P}(X)$ can be formulated as the following:
\begin{equation}
\mu_{\mathbb{P}_X}:=E_{X\sim \mathbb{P}_X}[\phi(X)]=E_{X\sim \mathbb{P}_X}[k(X,\cdot)],
\end{equation}
where $\mathbb{P}_X$ denotes $\mathbb{P}(X)$ for simplicity. If a kernel is characteristic, then the mean embedding $\mu_{\mathbb{P}_X}$ is injective. All the information about the distribution can be preserved \cite{sriperumbudur2010hilbert}. The kernel embedding cannot be computed directly and is usually estimated from observations. Given a sample $D = \{x_i\}_{i=1}^{n}$, where $n$ is the sample size of the domain, and the kernel embedding can be empirically estimated as the following:
\begin{equation}\label{empirical}
\hat{\mu}_{\mathbb{P}_X} = \frac{1}{n} \sum \limits_{i=1}^{n} \phi(x_i)=\frac{1}{n}\sum \limits_{i=1}^{n} k(x_i,\cdot).
\end{equation}

\subsection{Proposed Approach}
The proposed conditional invariant domain generalization (CIDG) method aims to find a conditional invariant representation $h(X)$ (a linear transformation of the original features) to reduce the variance of the conditional distribution $\mathbb{P}(h(X)|Y)$ across source domains. Suppose we can learn a perfect conditional invariant representation $h(X)$, which satisfies $\mathbb{P}^{s=i}(h(X)|Y)=\mathbb{P}^{s=j}(h(X)|Y)=\mathbb{P}^t(h(X)|Y)$, $i,j \in \{1,2,...,m\}$ and $\mathbb{P}^t$ denotes the target domain. We can gather all the source domains to construct a new single domain with a joint distribution $\mathbb{P}^t(h(X)|Y)\mathbb{P}^{\text{new}}(Y)$. Therefore, under the condition $\mathbb{P}^{\text{new}}(Y)=\mathbb{P}^t(Y)$, the learned $h(X)$ has the invariant joint distribution across training and test domains. Contrarily, the previous method can only guarantee that $\mathbb{P}(h(X))$ is invariant, and whether $\mathbb{P}(Y|h(X))$ is invariant remains unknown. If $\mathbb{P}^{\text{new}}(Y)$ is different from $\mathbb{P}^t(Y)$, our method cannot guarantee the invariance of the joint distribution either. Nevertheless, our method can at least guarantee invariant class-conditional distributions, which is still better than previous methods. This is because $\mathbb{P}(Y|h(X))$ is usually not very sensitive to the changes in the prior $\mathbb{P}(Y)$ if $h(X)$ is highly correlated with $Y$.

The learning of conditional invariant representations is achieved mainly through two regularization terms: total scatter of class-conditional distributions and scatter of class prior-normalized marginal distributions. The first term measures the variance of $\mathbb{P}(h(X)|Y)$ locally, while the second term measures the variance of $\mathbb{P}(h(X)|Y)$ globally. In addition to these two terms, we also incorporate several terms that measure the discriminative power of the representation $h(X)$ as done in the previous works. By minimizing the distribution variance across domains and maximizing the discriminative power in one objective function, we can obtain the conditional invariant representation which is predictable for the labels on unseen target domains. \par

\subsubsection{Total scatter of class-conditional distributions}
Suppose we have $m$ related domains $\{\mathbb{P}^1,\mathbb{P}^2,...,\mathbb{P}^m\}$ on $\mathcal{X} \times \mathcal{Y}$. The marginal distribution on $\mathcal{X}$ of the $s$-th domain is denoted as $\mathbb{P}^s_X$. Suppose the class labels of each domain vary from $1$ to $C$. For simplicity, the $j$-th class conditional distribution $\mathbb{P}^s(X|Y=j)$ of the $s$-th domain is denoted as $\mathbb{P}_j^s$. 
The total scatter of class-conditional distributions across domains can be formulated as:
\begin{equation}
\Psi\left(\{\mu_{\mathbb{P}_1^1}, \mu_{\mathbb{P}_2^1},...,\mu_{\mathbb{P}_C^m} \}\right) = \sum \limits_{j=1}^{C}\frac{1}{m} \sum \limits_{s=1}^m  \| \mu_{\mathbb{P}_j^s} - \overline{\mu}_j \|_{\mathcal{H}}^2, 
\end{equation}
where $\overline{\mu}_j = \frac{1}{m} \sum \limits_{s=1}^m \mu_{\mathbb{P}_j^s}$ and $\frac{1}{m}\sum \limits_{s=1}^m  \| \mu_{\mathbb{P}_j^s} - \overline{\mu}_j \|_{\mathcal{H}}^2$ is called the domain scatter \cite{ghifary8} or distributional variance \cite{muandet9}. Instead of measuring the domain scatter w.r.t. the marginal distributions $\mathbb{P}^s_X$ as done in previous works like \cite{ghifary8}, we measure the domain scatter w.r.t. each class-conditional distribution and then sum them together. 

Before introducing the computation of the above scatter, we first give the formulation of the learned feature transformation. Denote the feature matrix $\bm{X} = [x_1, x_2,...,x_{n}]^{\top} \in \mathbb{R}^{n \times d}$ as the data matrix of samples from $m$ source domains, where $d$ is the dimension of the feature space $\mathcal{X}$ and $n= \sum \nolimits_{s=1}^m n^s$. Define a set of functions $\bm{\Phi}=[\phi(x_1),\phi(x_2)...,\phi(x_n)]^{\top}$ related to the feature map $\phi: \mathbb{R}^d \rightarrow \mathcal{H}$. We aim to find a linear feature transformation $\bm{W}$ transforming $\mathcal{H}$ into a finite subspace $: \mathcal{H} \rightarrow \mathbb{R}^q$, that is $h(x)=\bm{W}^\top\phi(x)$.  According to the kernel principal component analysis (KPCA) \cite{scholkopf1998nonlinear}, the linear transformation can be formulated as the linear combination of $\bm{\Phi}, i.e., \bm{W} = \bm{\Phi}^{\top}\bm{B}$, where $\bm{B} \in \mathbb{R}^{n\times q}$ is the coefficient matrix. By using this representation, we can avoid explicitly computing the feature map $\phi$ and use the kernel trick instead.\par

For simplicity, denote $\Psi\left(\{\mu_{\mathbb{P}_1^1}, \mu_{\mathbb{P}_2^1},...,\mu_{\mathbb{P}_C^m} \}\right)$ as $\Psi^{con}$,
\begin{equation}
\begin{aligned}
\Psi^{con} &= \frac{1}{m} \sum \limits_{s=1}^m \sum \limits_{j=1}^{C} \| \mu_{\mathbb{P}_j^s} - \overline{\mu}_j \|_{\mathcal{H}}^2 \\
&= \frac{1}{m} \sum \limits_{s=1}^m \sum \limits_{j=1}^{C} Tr\left((\mu_{\mathbb{P}_j^s} -\overline{\mu}_j  )(\mu_{\mathbb{P}_j^s} -\overline{\mu}_j )^{\top}\right) \\
&= Tr\left(\frac{1}{m} \sum \limits_{s=1}^m \sum \limits_{j=1}^{C}(\mu_{\mathbb{P}_j^s} -\overline{\mu}_j  )(\mu_{\mathbb{P}_j^s} -\overline{\mu}_j )^{\top} \right),
\end{aligned}
\end{equation}
where $Tr(\cdot)$ is trace operator. To measure the distribution scatter of the distributions of $\mathbb{P}(h(X)|Y)$, we apply the linear feature transformation $\bm{W}$ to the above scatter and obtain
\begin{flalign}\label{con}
& \Psi_{\bm{B}}^{con} \nonumber\\
&= Tr\left(\frac{1}{m} \sum \limits_{s=1}^m \sum \limits_{j=1}^{C}\bm{B}^{\top}\bm{\Phi}(\mu_{\mathbb{P}_j^s} -\overline{\mu}_j  )(\mu_{\mathbb{P}_j^s} -\overline{\mu}_j )^{\top} \bm{\Phi}^{\top}\bm{B} \right) \nonumber\\
&= Tr\left(\bm{B}^{\top}\bm{H}\bm{B}\right),
\end{flalign}
where $\bm{H}$ is:
\begin{equation}\label{eq:H}
\bm{H} = \sum \limits_{s=1}^m \frac{1}{m}\sum \limits_{j=1}^{C} \bm{\Phi}(\mu_{\mathbb{P}_j^s} -\overline{\mu}_j  )(\mu_{\mathbb{P}_j^s} -\overline{\mu}_j )^{\top} \bm{\Phi}^{\top},
\end{equation}
in which $\mu_{\mathbb{P}_j^i}$ and $\overline{\mu}_j$ can be computed according to the empirical estimation shown in equation (\ref{empirical}). Denote $x_{k \sim j}^{s}$ as the $k$-th sample belonging to the $j$-th class in the $s$-th domain, where $s \in \{1, 2,...,m\}$ and $j \in \{1,2,...,C\}$. Let $n_j^s$ denote the sample size of the $j$-th class from the $s$-th domain, we have:
\begin{equation}
\begin{aligned}
\hat{\mu}_{\mathbb{P}_j^{s}} &= \frac{1}{n^s_j} \sum \limits_{k=1}^{n^s_j} \phi(x^s_{k \sim j}), ~
\hat{\overline{\mu}}_j = \frac{1}{m} \sum \limits_{s=1}^{m} \hat{\mu}_{\mathbb{P}_j^{s}},
\end{aligned}
\end{equation}
where $k\sim j$ denotes the indicies of examples in the $j$-th class.
\subsubsection{Scatter of class prior-normalized marginal distributions}
The scatter of each class-conditional distribution is estimated locally using the samples from that class. When the number of examples in each class is small, optimizing (\ref{con}) can easily overfit the data. To further improve the estimation accuracy, we propose another regularization term which measures the scatter of class-prior normalized marginal distributions. The new regularization term is able to measure the global distance between all class-conditionals. In the $s$-th domain, the marginal distribution is defined as 
\begin{equation}
\begin{aligned}
\mathbb{P}^s(X) = \sum_{j=1}^C \mathbb{P}^s(X|Y=j)\mathbb{P}^s(Y=j).
\end{aligned}
\end{equation}
If the class prior distribution $\mathbb{P}(Y)$ does not change across domains, and we can also find a feature representation that has an invariant class-conditional $\mathbb{P}(h(X)|Y)$ across source domains, we can say that $\mathbb{P}(h(X))$ is also domain-invariant, but not vice versa. Nevertheless, searching for a representation that reduces the discrepancy between the marginal distributions can to some extent reduce the discrepancy of class conditional distributions, though the original purpose was to match marginal distributions only \cite{muandet9,ghifary8}. However, if the class prior changes across source domains, the above statements are no longer true. That is to say, even if the class conditionals are domain-invariant, the marginal distribution are not invariant because of the changes in $\mathbb{P}(Y)$. To mitigate this issue, we propose to match the class-prior normalized marginal distribution, which is defined as follows:
\begin{eqnarray}
&\mathbb{P}^s_N(X) = \sum_{j=1}^C \mathbb{P}^s(X|Y=j)\frac{1}{C}
\end{eqnarray}
It can be seen that the class-prior normalized marginal distribution enforces the same prior probability for each class. Therefore, the changes in the prior distribution across source domains are adjusted, which guarantees that the prior-normalized marginal distribution is domain-invariant when the class conditionals are invariant. By embedding the class prior-normalzied marginal distribution into a Hilbert space, the scatter of the normalized marginal distribution across domains can be formulated as:
\begin{equation} \label{prior}
\begin{aligned}
\Psi^{prior} &= \frac{1}{m} \sum \limits_{s=1}^{m} \| \overline{\mu}_N - \mu_{\mathbb{P}^s_{N}} \|_{\mathcal{H}}^2,
\end{aligned}
\end{equation}
where $\mu_{\mathbb{P}_N^s} = E_{x\sim \mathbb{P}^s_{N}}[\phi(x)]$, and $\mathbb{P}^s_N$ is the prior-normalized marginal distribution of the $s$-th domain. $\overline{\mu}_N = \frac{1}{m} \sum \nolimits_{s=1}^m \mu_{\mathbb{P}^s_N}$ is the kernel mean of the class prior-normalized marginal distribution $\mathbb{P}_{N}$ of all domains. To learn the domain-invariant representation, we apply the linear feature transformation $\bm{W}$ to the above scatter, resulting in:
\begin{flalign}\label{prior1}
&\Psi^{prior}_{\bm{B}} \nonumber\\
&= Tr\left(\frac{1}{m} \sum \limits_{s=1}^m \bm{B}^{\top} \bm{\Phi} (\overline{\mu}_N - \mu_{\mathbb{P}_N^s})(\overline{\mu}_N - \mu_{\mathbb{P}_N^s})^{\top} \bm{\Phi}^{\top} \bm{B} \right)\nonumber \\
&= Tr\left(\bm{B}^{\top}\bm{L}\bm{B}\right),
\end{flalign}
where $\bm{L}$ can be formulated as follows:
\begin{equation}\label{eq:L}
\bm{L} = \frac{1}{m} \sum \limits_{s=1}^{m} \bm{\Phi}(\overline{\mu}_N - \mu_{\mathbb{P}_N^s})(\overline{\mu}_N - \mu_{\mathbb{P}_N^s})^{\top} \bm{\Phi}^{\top}.
\end{equation}
$\mu_{\mathbb{P}_N^s}$ in (\ref{eq:L}) can be empirically estimated from the observations as:
\begin{equation}
\hat{\mu}_{\mathbb{P}_N^s} = \frac{1}{C} \sum \limits_{j=1}^{C} \frac{1}{n^s_j} \sum \limits_{k=1}^{n^s_j} \phi(x_{k \sim j}^{s}).
\end{equation}
Note that if $n_j^s$ are identical for all $j$, that is the classes are balanced, the class prior-normalized marginal distribution reduces to the empirical estimate of the original marginal distribution $\hat{\mu}_{\mathbb{P}^s} = \frac{1}{n^s} \sum \limits_{k=1}^{n^s} \phi(x_{k}^{s})$ adopted in \cite{muandet9,ghifary8}.

\subsubsection{Preserving Discriminative Power}
In addition to the above proposed two domain-invariance regularization terms, we also consider extra terms to preserve the discriminativeness of the learned representation. There have been plenty of works in supervised dimension reduction in the $i.i.d.$ case, and kernel Fisher discriminant analysis \cite{mika1999fisher} is a representative method which has been used in domain generalization \cite{ghifary8}. Becasue the focus of our method is to better learn the domain-invariant representations, we incorporate kernel Fisher discriminant analysis for fair comparison to existing methods. Specifically, the examples with the same label should be similar and the examples with different labels should be well separated. These two constraints can be formulated as two regularization terms: within-class scatter and between-class scatter, which are briefly described as follows. \par
Between-class scatter:
\begin{equation}\label{between}
\Psi^{between}_{\bm{B}} = Tr(\bm{B}^{\top}\bm{P}\bm{B}),
\end{equation}
where matrix $\bm{P}$ can be computed as: 
\begin{equation}{eq:P}
\bm{P} = \sum \limits_{j=1}^C n_j \bm{\Phi} (\mu_j - \overline{\mu}_b)(\mu_j - \overline{\mu}_b)^{\top} \bm{\Phi}^{\top},
\end{equation}
and $n_j= \sum \nolimits_{s=1}^m n^s_j$ denotes the number of examples in the $j$-th class from all domains. Note that $\mu_j$ and $\overline{\mu}_b$ can be empirically estimated as $\hat{\mu}_j = \frac{1}{n_j} \sum \nolimits_{s=1}^{m} \sum \nolimits_{k=1}^{n_j^s} \phi(x^s_{k\sim j})$ and $\hat{\overline{\mu}}_b = \frac{1}{n} \sum \nolimits_{j=1}^C n_j \hat{\mu}_j$. \par

Within-class scatter:
\begin{equation}\label{within}
\Psi^{within}_{\bm{B}} = Tr(\bm{B}^{\top}\bm{Q}\bm{B}),
\end{equation}
where the matrix $\bm{Q}$ can be computed as:
\begin{equation}\label{eq:Q}
\bm{Q} = \sum \limits_{j=1}^C \sum \limits_{s=1}^m \sum \limits_{k=1}^{n_j^s} \bm{\Phi}(\phi(x_{k\sim j}^s) - \mu_j)(\phi(x_{k\sim j}^s) - \mu_j)^{\top} \bm{\Phi}^{\top}.
\end{equation}

\subsubsection{Objective Function and Optimization}
In this subsection, we first formulate our objective function with the above regularization terms and then find the solutions by maximizing the objective function. \par

The proposed CIDG aims to learn an invariant feature transformation by solving the following optimization problem:
\begin{equation}
\argmax_{\bm{B}} \frac{\Psi^{between}_{\bm{B}}}{\Psi^{con}_{\bm{B}} + \Psi^{prior}_{\bm{B}} + \Psi^{within}_{\bm{B}}}.
\end{equation}
The numerator enforces the distance between features in different classes to be large. The denominator aims to learn a conditional invariant feature representation and reduce the distance between features in the same class simultaneously.\par

Replace the scatters with equation, (\ref{con}), (\ref{prior1}), (\ref{between}), (\ref{within}) and introduce several trade-off parameters $\gamma, \alpha$, the above objective function can be reformulated as follows:
\begin{equation}\label{obj_1}
\argmax_{\bm{B}} \frac{Tr(\bm{B}^{\top}\bm{P}\bm{B})}{Tr(\bm{B}^{\top}(\gamma\bm{H} + \alpha\bm{L} + \bm{Q})\bm{B})},
\end{equation}
where $0<\gamma$, $0<\alpha$ are trade-off parameters, which need to be selected according to the validation set. \par

Note that the above objective function is invariant when rescaling $\bm{B} \rightarrow \eta \bm{B}$, where $\eta$ is a constant. Consequently, (\ref{obj_1}) can be reformulated as the following constrained optimization problem: 
\begin{equation}
\begin{aligned}
\argmax_{\bm{B}} \quad& Tr(\bm{B}^{\top}\bm{P}\bm{B}) \\
s.t. \quad & Tr(\bm{B}^{\top}(\gamma\bm{H} + \alpha\bm{L} + \bm{Q})\bm{B}) = 1,
\end{aligned}
\end{equation}
which yields Lagrangian:
\begin{equation}\label{lagrange}
\begin{aligned}
L(\bm{B}) = & Tr(\bm{B}^{\top}\bm{P}\bm{B}) \\
 &- Tr((\bm{B}^{\top}(\gamma\bm{H} + \alpha\bm{L} + \bm{Q})\bm{B} - \bm{I}_q) \bm{\Gamma}),
\end{aligned}
\end{equation}
where $\bm{I}_q$ is an identity matrix of dimension $q$ and $\Gamma = diag(\lambda_1,\lambda_2,...,\lambda_q)$ is a diagonal matrix with the Lagrange multipliers aligned in the diagonal. Solving (\ref{lagrange}) by setting the derivative w.r.t. $\bm{B}$ to be zero, we arrive at a standard eigenvalue decomposition problem:
\begin{equation}\label{result}
\bm{P}\bm{B} = (\gamma \bm{H} + \alpha \bm{L} + \bm{Q})\bm{B}\bm{\Gamma}.
\end{equation}
In practice, the term $(\gamma \bm{H} + \alpha \bm{L} + \bm{Q})$ is added by a small constant $\epsilon \bm{I}$ to get a more stable solution, becoming $(\gamma \bm{H} + \alpha \bm{L} + \bm{Q} + \epsilon \bm{I})$. We summarize the algorithm of our CIDG in Algorithm \ref{algorithm}.

\begin{algorithm}
\caption{Conditional invariant domain generalization}
\label{algorithm}
\begin{algorithmic}[1]
\REQUIRE $m$ source domains with datasets $S_D = \{D_s = \{x_i^s,y_i^s\}_{i=1}^{n^s}, s = \{1,2,...,m\}\}$, trade-off parameters $\gamma,\alpha$.
\ENSURE Invariant feature transformation $\bm{B}^*$ and corresponding eigenvalues $\bm{\Gamma}^*$
\STATE Construct kernel matrix $\bm{K}$ from data samples of all domains, $\bm{K}(i,j) = k(x_i, x_j)$, $\forall x_i,x_j \in S_D$, and construct matrices $\bm{H},\bm{L},\bm{P},\bm{Q}$ from equations (\ref{between}), (\ref{con}), (\ref{prior1}), (\ref{within}).
\STATE Centering the kernel matrix $\bm{K} \leftarrow \bm{K} - \bm{1}_n\bm{K} - \bm{K}\bm{1}_n + \bm{1}_n\bm{K}\bm{1}_n$, where $n = \sum \nolimits_{i=1}^m n^s$ and $\bm{1}_n \in \mathbb{R}^{n \times n}$ denotes a matrix with all entries equal to $\frac{1}{n}$. 
\STATE Solve the equation (\ref{result}) to get the optimal feature transformation matrix $\bm{B}^*$ and the corresponding eigenvalues $\bm{\Gamma}^*$ with the first $q$ leading eigenvalues. 
\STATE When given a target domain with a set of data $D_t = \{x_i^t,y_i^t\}_{i=1}^{n^t}$, construct a kernel matrix $\bm{K}^t$ with samples from source domains and samples from the target domain, $\bm{K}^t(i,j) = k(x_i, x_j), \forall x_i \in S_D, x_j \in D_t$. Then we apply the centering operation to $\bm{K}^t \leftarrow \bm{K}^t - \bm{1}_n\bm{K}^t - \bm{K}^t\bm{1}_{n^t} + \bm{1}_n \bm{K}^t \bm{1}_{n^t}$, where $\bm{1}_{n^t} \in \mathbb{R}^{n^t \times n^t}$ denotes a matrix with all entries equal to $\frac{1}{n}$.
\STATE The learned feature matrix of the target domain can be computed as $\bm{X}^* = (\bm{K}^t)^{\top}\bm{B}^*(\bm{\Gamma}^*)^{-\frac{1}{2}}$.
\end{algorithmic}
\end{algorithm}

\begin{table*}
\begin{center}
\label{synthetic}
\begin{tabular}{|c|c|c|c|c|c|c|c|c|c|}
\hline
 domain index & \multicolumn{3}{c|}{domain 1} &  \multicolumn{3}{c|}{domain 2} &  \multicolumn{3}{c|}{domain 3}  \\
\hline
class index &$ 1$ & $2$ & $3$ & $ 1$ & $2$ & $3$ &$ 1$ & $2$ & $3$  \\
\hline
x &(1,0.3)& (2, 0.3) & (3, 0.3) & (3.5, 0.3) & (4.5, 0.3) & (5.5, 0.3) & (8, 0.3) & (9.5, 0.3) & (10, 0.3)    \\
\hline
y &(2,0.3)& (1, 0.3) & (2, 0.3) & (2.5, 0.3) & (1.5, 0.3) & (2.5, 0.3) & (2.5, 0.3) & (1.5, 0.3) & (2.5, 0.3)   \\
\hline
\# samples & 30 & 20 & 30 & 20 & 60 & 40 & 40 & 40  & 40\\
\hline
\end{tabular}
\caption{Details of the generated distributions of three domains.}
\end{center}
\end{table*}

\begin{figure*}
\centering
\label{visualization}
\includegraphics[width=0.8\textwidth]{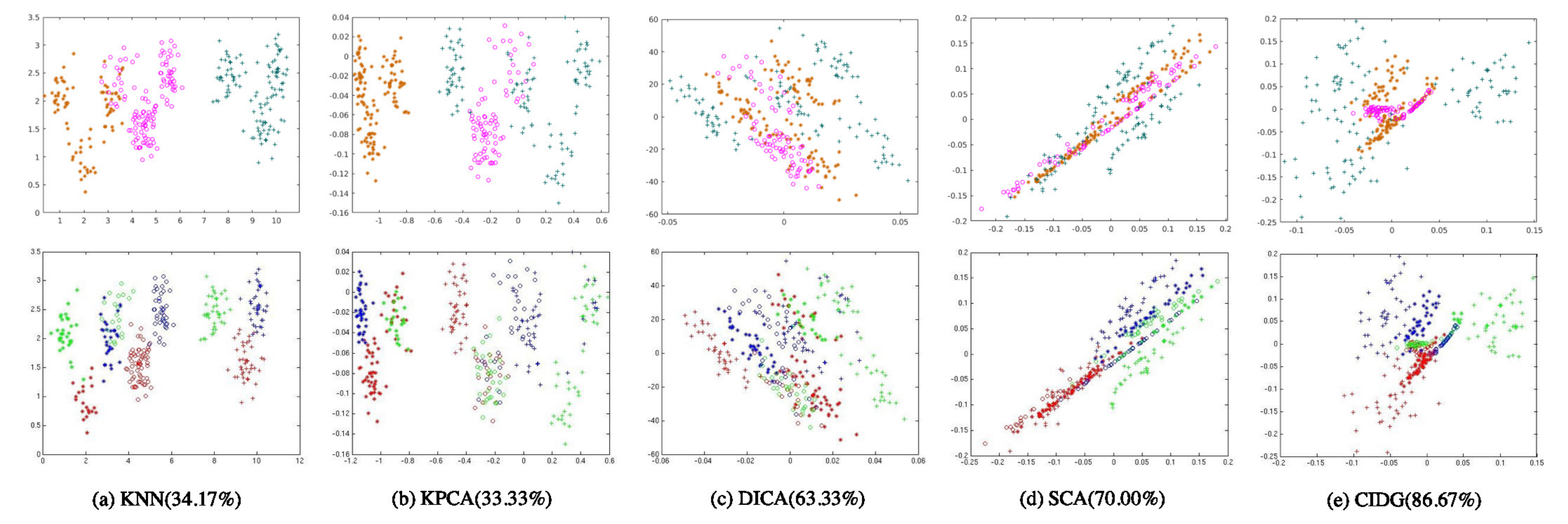}
\caption{Performance comparison between different methods. The figures in the first row visualize the samples according to three different domains (yellow, magenta, cyan). The figures in the second row visualize the samples of three classes (green, red, blue) in different domains (star, circle, cross). Note that the left two domains (yellow, magenta) are source domains and the right one (cyan) is target domain.}
\end{figure*}

\section{Experiments}
In this section, we conduct experiments on one synthetic data and two real-world image classification datasets to demonstrate the effectiveness of our conditional invariant domain generalization (CIDG) method. The synthetic data are two dimensional, which facilitate the comparison of the performance of different methods through the visualization of the data distribution. The two real-world image classification datasets are the VLCS and Office+Caltech datasets, which are widely used datasets to evaluate the performance of domain generalization and domain adaptation \cite{ghifary8,gong10,khosla2012undoing}. We compare our CIDG with several state-of-the-art domain generalization methods, which are summarized below.
\begin{itemize}
\item K-nearest neighbors (KNN) using the original features, which servers as the baseline method. 
\item Kernel principal component analysis (KPCA) \cite{scholkopf1998nonlinear} which finds the dominant components of the original features. KNN is applied for classification on the KPCA features.
\item Undo-Bias \cite{khosla2012undoing}, which is a multi-task learning method aims to reduce the data bias. Because undo-bias is a binary classification algorithm, we use the one-vs-rest strategy for multi-class classification. 
\item  Domain invariant component analysis (DICA) \cite{muandet9}, which is a domain generalization method learns an domain-invariant feature representation in terms of marginal distributions. We use KNN to do classification on the learned feature representation. 
\item Scatter component analysis (SCA) \cite{ghifary8}, which is a another method that learns domain-invariant features in terms of marginal distributions. The method incorporates discriminative terms and domain scatter terms into a unified framework.
\end{itemize}
Note that we have also conducted experiments using kernel finsher discriminant analysis (FDA), however, it performs worse than KPCA. Consequently, we do not report the results of KLDA in this paper.

\begin{table*}
\begin{center}
\label{VLCS}
\begin{tabular}{|c|c|c|c|c|c|c|c|}
\hline
 Source & Target & 1NN & KPCA & DICA & Undo-bias & SCA & CIDG \\
\hline
L,C,S & V & $53.27 \pm 1.52 $ & $58.62 \pm 1.44$ & $58.29 \pm 1.51$ & $57.73 \pm 1.02$ & $57.48 \pm 1.78$ & $\bm{65.65 \pm 0.52}$ \\
\hline
V,C,S & L & $50.35 \pm 0.94$ & $53.80 \pm 1.78$ & $50.35 \pm 1.45$ & $58.16 \pm 2.13$ & $52.07 \pm 0.86$ & $\bm{60.43 \pm 1.57}$  \\
\hline
V,L,S & C & $76.82 \pm 1.56$ & $85.84 \pm 1.64$ & $73.32 \pm 4.13$ & $82.18 \pm 1.77$ & $70.39 \pm 1.42$ & $\bm{91.12 \pm 1.62}$  \\
\hline
V,C,L & S & $51.78 \pm 2.07$ & $53.23 \pm 0.62$ & $54.97 \pm 0.61$ & $55.02 \pm 2.53$ & $54.46 \pm 2.71$ & $\bm{60.85 \pm 1.05}$  \\
\hline
C,S & V,L & $52.44 \pm 1.87$ & $55.74 \pm 1.01$ & $53.76 \pm 0.96$ & $56.83 \pm 0.67$ & $56.05 \pm 0.98$ & $\bm{59.25 \pm 1.21}$  \\
\hline
C,L & V,S & $45.04 \pm 2.49$ & $45.13 \pm 3.01$ & $44.81 \pm 1.62$ & $52.16 \pm 0.80$ & $48.97 \pm 1.04$ & $\bm{54.04 \pm 0.91}$  \\
\hline
C,V & L,S & $47.09 \pm 2.49$ & $55.79 \pm 1.57$ & $49.81 \pm 1.40$ & $59.00 \pm 2.49$ & $53.47 \pm 0.71$ & $\bm{61.61 \pm 0.67}$  \\
\hline
L,S & V,C & $57.09 \pm 1.43$ & $\bm{58.50 \pm 3.84}$ & $44.09 \pm 0.58$ & $51.16 \pm 3.52$ & $49.98 \pm 1.84$ & $55.65 \pm 3.57$  \\
\hline
L,V & S,C & $59.21 \pm 1.84$ & $63.88 \pm 0.36$ & $61.22 \pm 0.95$ & $64.26 \pm 2.77$ & $66.68 \pm 1.09$ & $\bm{70.89 \pm 1.31}$  \\
\hline
V,S & L,C & $58.39 \pm 0.78$ & $64.56 \pm 0.99$ & $60.68 \pm 1.36$ & $68.58 \pm 1.62$ & $63.29 \pm 1.34$ & $\bm{70.44 \pm 1.43}$  \\
\hline
\end{tabular}
\caption{Performance comparison between different methods with respect to accuracy ($\%$) on VLCS dataset.}
\end{center}
\end{table*}

\begin{table*}
\begin{center}
\label{office+caltech}
\begin{tabular}{|c|c|c|c|c|c|c|c|}
\hline
 Source & Target & 1NN & KPCA & DICA & Undo-bias & SCA & CIDG \\
\hline
W,D,C & A & $87.65 \pm 2.46$ & $90.92 \pm 1.03$ & $80.34 \pm 2.65$ & $89.56 \pm 1.55$ & $89.97 \pm 1.85$ & $\bm{93.24 \pm 0.71}$ \\
\hline
A,W,D & C & $67.00 \pm 0.67$ & $74.23 \pm 1.34$ & $64.55 \pm 2.85$ & $82.27 \pm 1.49$ & $77.90 \pm 1.28$ & $\bm{85.07 \pm 0.93}$  \\
\hline
A,W,C & D & $97.36 \pm 1.92$ & $94.34 \pm 1.19$ & $93.21 \pm 1.92$ & $95.28 \pm 2.45$ & $93.21 \pm 3.50$ & $\bm{97.36 \pm 0.92}$  \\
\hline
A,C,D & W & $82.11 \pm 0.67$ & $88.84 \pm 2.17$ & $69.68 \pm 3.22$ & $90.18 \pm 2.10$ & $81.26 \pm 3.15$ & $\bm{90.53 \pm 2.66}$  \\
\hline
A,C & D,W & $60.95 \pm 1.31$ & $75.81 \pm 2.94$ & $60.41 \pm 1.94$ & $80.24 \pm 2.21$ & $76.89 \pm 0.99$ & $\bm{83.65 \pm 2.24}$  \\
\hline
D,W & A,C & $60.47 \pm 0.99$ & $65.75 \pm 1.74$ & $43.02 \pm 3.24$ & $\bm{74.14 \pm 3.45}$ & $69.53 \pm 1.87$ & $65.91 \pm 1.42$  \\
\hline
A,W & C,D & $71.11 \pm 0.81$ & $76.26 \pm 1.13$ & $69.29 \pm 1.77$ & $81.77 \pm 1.77$ & $78.99 \pm 1.54$ & $\bm{83.89 \pm 2.97}$  \\
\hline
A,D & C,W & $60.95 \pm 1.31$ & $75.81 \pm 2.94$ & $68.49 \pm 2.88$ & $81.23 \pm 2.17$ & $75.84 \pm 1.66$ & $\bm{84.66 \pm 3.27}$  \\
\hline
C,W & A,D & $89.08 \pm 2.26$ & $91.45 \pm 1.27$ & $83.01 \pm 2.42$ & $91.73 \pm 0.67$ & $90.46 \pm 1.72$ & $\bm{93.41 \pm 0.92}$  \\
\hline
C,D & A,W & $86.19 \pm 1.58$ & $90.36 \pm 1.26$ & $79.69 \pm 1.11$ & $90.67 \pm 1.87$ & $88.61 \pm 0.38$ & $\bm{91.70 \pm 1.35}$  \\
\hline
\end{tabular}
\caption{Performance comparison between different methods with respect to accuracy ($\%$) on office+caltech dataset.}
\end{center}
\end{table*}

\subsection{Synthetic Dataset}
In this section, we randomly generate two dimensional examples for source domains and target domain from different Gaussian distributions $\mathcal{N}(\mu, \sigma)$, where $\mu$ is the mean and $\sigma$ is the standard deviation. The values of mean $\mu$ and standard deviation $\sigma$ pairs $(\mu, \sigma)$  of different classes in three domains are shown in Table \ref{synthetic}. We consider the first two domains as source domains and the third one as a target domain. The first row of Figure \ref{visualization} visualizes the samples from three different domains corresponding to three different colors (yellow, magenta, cyan), and the domains are domain $1$, domain $2$ and domain $3$ from left to right. The second row of Figure \ref{visualization} shows that each domain has three clusters (green, red, blue) corresponding to three different classes and the domains are represented by different shapes (star, circle, cross). The first column illustrates the raw feature distributions. \par

We compare our CIDG with KNN, KPCA, DICA, and SCA to evaluate the distributions of the learned feature representation across domains. Since Undo-Bias is a SVM-based method that does not need to explicitly learn a feature representation, we do not compare the results with Undo-Bias on synthetic data. We use the RBF kernel for all the methods involving computation of kernel matrices. In all experiments, domain 1 and domain 2 are used as source domains and domain 3 is used as the unseen target domain. From the results in Figure \ref{visualization}, we can see that the proposed CIDG achieves the best accuracy of $86.67\%$. KPCA almost has no improvement over the baseline KNN method on the synthetic dataset. DICA can cluster one class (blue) well but performs badly for the other two classes. SCA can learn better feature distribution but the blue class and the green class are mixed in the learned representation. Additionally, the samples in the same class lie in a line rather than reside in a clear cluster. Our CIDG can learn more robust feature representations and the learned features in the same class are distributed in a well-shaped cluster. \par

\subsection{VLCS Dataset}
VLCS is an image classification dataset widely used for evaluating the performance of domain generalization. This dataset contains images from four different sub-datasets corresponding to four domains: PASCAL VOC2007 (V) \cite{everingham2010pascal}, LabelMe (L) \cite{russell2008labelme}, Caltech-101 (C) \cite{griffin2007caltech}, and SUN09 (S) \cite{choi2010exploiting}. Five shared classes (bird, car, chair, dog and person) are selected from these four datasets. The images are preprocessed by subtracting the mean values and cropped on the central $224 \times 224$ region out of the $256 \times 256$ resized images. Then the preprocessed images are fed into the DeCAF network and extracted the 4096 dimensional DeCAF\textsubscript{6} features \cite{donahue2014decaf}. 
We randomly select $70\%$ of the data as training set from each domain and repeat the random selection five times.  The mean classification accuracy and standard deviation of the five random selection are given for each method. All parameters are selected through validation, in which $30\%$ of the training data is selected as validation set.  All kernel methods use a RBF kernel and the learned features are classified using KNN except for Undo-Bias. The results are shown in Table \ref{VLCS}. \par

From the results in Table \ref{VLCS}, we can see that our conditional invariant domain generalization (CIDG) performs the best on 9 of the 10 domain generalization tasks. KPCA performs the best when L,S are source domains and V,C are target domains. Note that almost all the domain generalization methods outperform the 1NN on raw features. However, some methods on several domain tasks perform even worse than 1NN on raw features. This is mainly because that features of real world images are complicated and noisy.  The learned features are not discriminative when generalized to target domains. \par

\subsection{Office+Caltech Dataset}
The Office+Caltech image dataset consists of ten overlapping categories between the Office dataset and the Caltech-256 dataset (C). Because the Office dataset contains three sub-datasets: AMAZON (A), DSLR (D), and WEBCAM (W), we have four different domains in total. Similarly, We randomly select $70\%$ of the data as training set from each domain and repeat the random selection five times. The mean classification accuracy and standard deviation of the five random selection are reported for each method. The feature extraction is the same as that used for the VLCS dataset except we use the CAFFE network \cite{jia2014caffe} instead of the DeCAF network. The other settings are the same as those in experiments on the VLCS dataset. \par

From the results in Table \ref{office+caltech}, we can find that the proposed CIDG achieves the best performance on 9 of the 10 domain generalization tasks. This further validates that enforcing conditional invariance is more reasonable than enforcing only marginal invariance. Note that Undo-bias is a SVM-based method. It is possibly the main reason why it outperforms CIDG when using D,W as source domains and A,C as target domains. \par

\section{Conclusion}
In this paper, we have proposed a conditional invariant domain generalization approach considering the situation that both $\mathbb{P}(X)$ and $\mathbb{P}(Y|X)$ change across domains. Different from previous works which assume that only $\mathbb{P}(X)$ changes, our proposed method can learn representations that have invariant joint distribution $\mathbb{P}(h(X),Y)$ across domains if the prior distribution $\mathbb{P}(Y)$ does not change between the source domains and the target domains. Two regularization terms that enforce class-conditional distribution invariance across domains are proposed and validated on both synthetic and real datasets. 
\section*{Acknowledgments}
This work was supported by National Key Research and Development Program of China 2017YFB1002203, NSFC No.61572451, No.61390514, and No. 61632019, Youth Innovation Promotion Association CAS CX2100060016, Fok Ying Tung Education Foundation WF2100060004, and Australian Research Council Projects FL-170100117, DP-180103424, DP-140102164, LP-150100671. 

\bibliographystyle{aaai}
\bibliography{aaai}
\end{document}